\documentclass{article}
\usepackage{spconf,amsmath,graphicx,amstext,amssymb,epsf,epstopdf}

\title{Improved Eigenfeature Regularization for Face Identification}
\name{Bappaditya Mandal}
\address{Email: bmandal@i2r.a-star.edu.sg\\Institute for Infocomm Research, A*STAR, Singapore}
\begin{document}
%
\maketitle
\begin{abstract}
In this work, we propose to divide each class (a person) into subclasses using spatial partition trees which helps in better capturing the intra-personal variances arising from the appearances of the same individual. We perform a comprehensive analysis on within-class and within-subclass eigenspectrums of face images and propose a novel method of eigenspectrum modeling which extracts discriminative features of faces from both within-subclass and total or between-subclass scatter matrices. Effective low-dimensional face discriminative features are extracted for face recognition (FR) after performing discriminant evaluation in the entire eigenspace. Experimental results on popular face databases (AR, FERET) and the challenging unconstrained YouTube Face database show the superiority of our proposed approach on all three databases.
\end{abstract}
\begin{keywords}
Feature extraction, discriminant analysis, subspace learning, face identification.
\end{keywords}

\section{Introduction}
In multi-class classification, Fisher-Rao's linear discriminant analysis (LDA) minimizes Bayesian error when sample vectors of each class are generated from multivariate Normal distributions of same covariance matrix but different means (homoscedastic data \cite{Duda,Mandal6}). However, for real-world face images, the classes have random (or heteroscedastic \cite{Duda}) distributions, the variances are quite large and the data are of very high dimensionality. So the estimation of variances (using within-class and between-class scatter matrices in LDA) are limited to averages of all the possible variations among  training samples. This limits the usages of LDA in face image data for FR, where large number of classes with high dimensional data are involved \cite{Cevikalp,Chen1,Jiang2}. Mixture discriminant analysis (MDA) \cite{Hastie1} models each class as a mixture of Gaussians, rather than a single Gaussian as in LDA. MDA uses a clustering procedure to find subclass partitions of the data and then incorporate this information into the LDA criterion.

Subclass discriminant analysis (SDA) in \cite{Zhu1} maximizes the distances between both the means of classes and the means of the subclasses. SDA emphasizes the role of class separability rather than the discriminant information in the within-subclass scatter matrix, hence it may not capture the crucial discriminant information in the within-subclass variances for FR \cite{Liu}. Mixture subclass discriminant analysis (MSDA), an improvement over SDA is presented in \cite{Gkalelis1}. In this approach, a subclass partitioning procedure along with a non-Gaussian criterion are used to derive the subclass division that optimizes the MSDA criterion, this has been extended to fractional MSDA and kernel MSDA in \cite{Gkalelis2}.

All the above approaches discard the null space of either within-class and within-subclass scatter matrices, which plays a very crucial role in the discriminant analysis of faces \cite{Liu,Chen1,Cevikalp,Mandal3}. Xudong \emph{et al.} \cite{Jiang4,Jiang5} proposed an eigenfeature regularization method (ERE) which partitions the eigenspace into various subspaces. However, their variances are extracted from within-class scatter matrix and does not consider partitions within each class. Hence, this method would fail in capturing the crucial within-subclass discriminant information, even for part-based recognition \cite{Mandal7}.

Another class of emerging algorithms is the deep learning which uses convolutional neural network and millions of (external) face images for training and obtain very high accuracy rates \cite{Taigman1,Yisun1}. However, our proposed method is still attractive because it uses small number of training samples and does not use any external training data but can achieve comparable performances and is suitable for mobile devices \cite{Mandal5,Mandal10}.

\section{Discriminant Analysis}

\subsection{Scatter matrices and discriminant analysis}
The problem of discriminant analysis is generally solved by maximization of the Fisher criterion \cite{Duda,Mandal8}. This involves between-class ($S_b$) and within-class
($S_w =\frac{1}{n}\sum^C_{i=1}\sum^{n_i}_{j=1}(x_{ij}-\mu_i)(x_{ij}-\mu_i)^T$) scatter matrices, where $C$ is the number of classes or persons, $\mu_i$ is the sample mean of class $i$, $\mu$ is the global mean, $x_{ij}$ $\in \mathbb{R}^l$, by lexicographic ordering the pixel elements of image of size $l=width \times height$, is the $j^{th}$ sample of class $i$, $n_i$ is the number of samples in $i^{th}$ class and $n=\sum_{i=1}^Cn_i$ is the total number of samples.

LDA assumes that the class distributions are homoscedastic, which is rarely true in practice for FR. We assume that there exist subclass homoscedastic partitions of the data and model each class as mixtures of Gaussians \cite{Lu8} subclasses, whose objective function is defined as
$J(\Psi)=\frac{tr(\Psi^TS_{bs}\Psi)}{tr(\Psi^TS_{ws}\Psi)},$
where $tr$ represents trace of a matrix, $\Psi$ denotes a transformation matrix, $S_{bs}$ is the between-subclass scatter matrix and $S_{ws}$ is the within-subclass scatter matrix defined as
\begin{equation}\label{eqn_3_Sws}
S_{ws}=\sum^C_{i=1}p_i\sum^{H_i}_{j=1}\frac{q_{_{H_i}}}{G_{ij}}\sum_{k=1}^{G_{ij}}(x_{ijk}-\mu_{ij})(x_{ijk}-\mu_{ij})^T.
\end{equation}
$H_i$ denotes the number of subclasses of the $i^{th}$ class and $G_{ij}$ denotes the number of samples in $j^{th}$ subclass of $i^{th}$ class. $x_{ijk}$ $\in \mathbb{R}^{l}$ is the $k^{th}$ image vector in $j^{th}$ subclass of $i^{th}$ class. $\mu_{ij}=\frac{1}{G_{ij}}\sum_{k=1}^{G_{ij}}x_{ijk}$ is the sample mean of $j^{th}$ subclass of the $i^{th}$ class. $p_i$ and $q_{_{H_i}}$ are the estimated prior probabilities. If we assume that each class and subclasses have equal prior probabilities then $p_i=\frac{1}{C}$ and $q_{_{H_i}}=\frac{1}{H_i}$. If $S_{ws}$ is nonsingular, the optimal projection vectors $\Psi$ is chosen as the matrix with orthonormal columns which maximizes the ratio of the determinant of the between-subclass matrix of the projected samples to the determinant of the within-subclass scatter of the projected samples.

\subsection{Partitioning of a face class into subclasses} \label{ssec:pt}
We investigate various popular spatial partition trees to partition each face class into subclasses for within-subclass discriminant analysis, \cite{freund2007learning,Wang4}: (i) $k$-d tree, (ii) RP tree, (iii) PCA tree, and (iv) $k$-means tree. $k$-d trees and RP trees are built by recursive binary splits. They differ only in the nature of the split. Unlike $k$-d tree, RP tree adapts to intrinsic low dimensional structure without having to explicitly learn face structure. In PCA tree, the partition axis is obtained by computing the principal eigenvector of the covariance matrix of the face image data. Since face image appear very differently under various contexts this kind of partitioning would be advantageous for data that are heterogeneously distributed in all dimensions. $k$-means tree is built based on nearest neighbor (NN) clustering of face appearances. In our experiments, we use the implementations of spatial partitioning trees by Freund \emph{et al.} \cite{freund2007learning}, with their default parameters of maximum depth up to eight layers and no overlap in samples splitting.

\subsection{Regularization of within-subclass eigenspace} \label{ssec:rs}

\subsubsection{Divisions in within-subclass eigenspace}
Formation of subclasses using spatial partition tree helps in capturing the variances more closely in appearances of the same individual. We compute the eigenvectors $\Psi^{ws}=\{\psi_1^{ws},\ldots,\psi_l^{ws}\}$ corresponding to the eigenvalues $\Lambda^{ws}=\{\lambda_1^{ws},\ldots,\lambda_l^{ws}\}$ of $S_{ws}$ described by (\ref{eqn_3_Sws}), where the eigenvalues are sorted in descending order.
\begin{figure}[!htp]
\centering
\begin{minipage}[b]{5.75\linewidth}
\includegraphics*[width=1.6in]{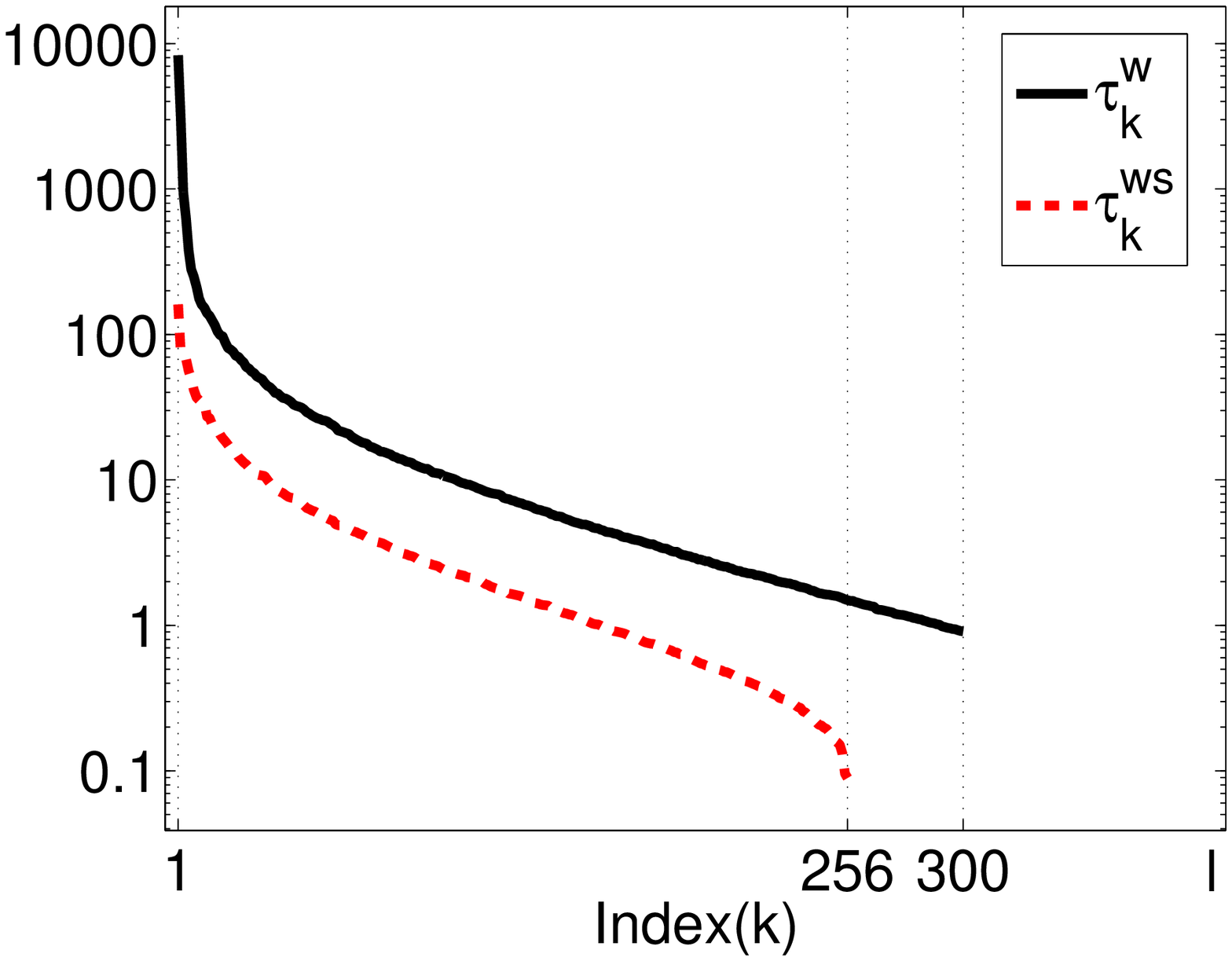}
\hspace{-0.3cm} \includegraphics*[trim=60 0 0 0,clip,width=1.75in]{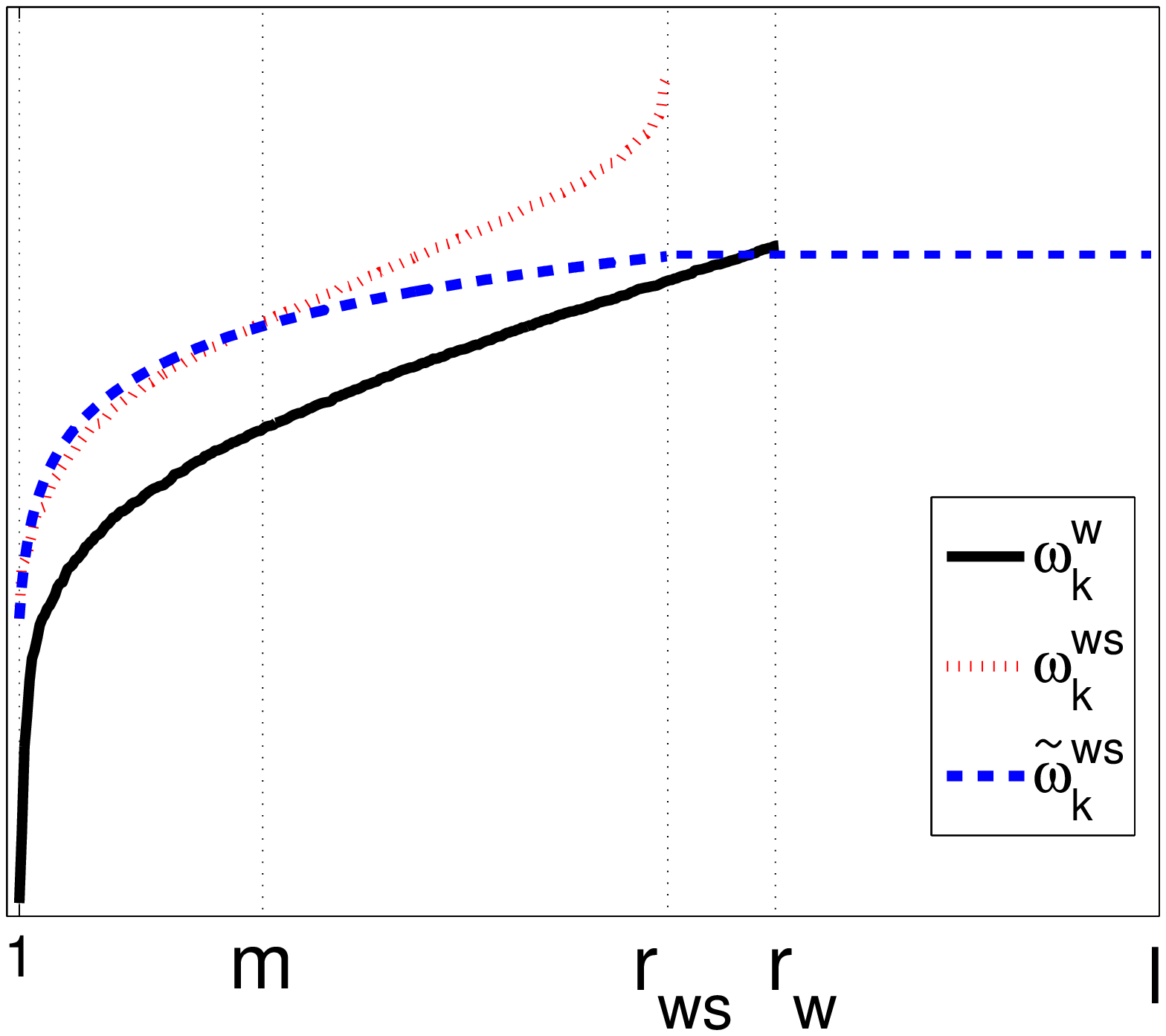}
\end{minipage}
\caption{Left, A typical real eigenspectrum computed from within-class and within-subclass sorted in descending order. Right, Inverse of a typical real eigenspectrum computed from within-class and within-subclass and weighting function $\tilde{\omega}^{ws}_k$.}
\label{fig1_eigenValuesSwSws}
\end{figure}

A typical plot of $\tau^{w}_k=\sqrt{\lambda^{w}_k}$ and $\tau^{ws}_k=\sqrt{\lambda^{ws}_k}$ computed from within-class ($\Lambda^{w}$) and within-subclass ($\Lambda^{ws}$) scatter matrices on real face training images are shown in Fig. \ref{fig1_eigenValuesSwSws} left. The decay of the eigenvalues of $S_{ws}$ is much faster than $S_{w}$. This is because within-class matrices capture variances that arise in the face appearances of the same individual. Forming subclasses, further lower their values. Their differences are of the order $10^2$ for larger and smaller ones. In both curves, as pointed out in \cite{Friedman,Mandal4}, the characteristic bias is most pronounced when the population eigenvalues tend toward equality, and it is correspondingly less severe when their values are highly disparate. Therefore, the smallest eigenvalues are biased much more than the largest ones \cite{Friedman,Jiang4,Mandal4}.

The whitened eigenvector matrix $\overline{\Psi}^{ws}=\{\psi_1^{ws}/\tau^{ws}_1,\ldots,\\\psi_l^{ws}/\tau^{ws}_l\}$, $\tau^{ws}_k=\sqrt{\lambda^{ws}_k}$ in Fig. \ref{fig1_eigenValuesSwSws} left, is used to project the image vector $x_{ij}$ before constructing the between-subclass scatter matrix. This is equivalent to image vector $x_{ij}$ first transformed by the eigenvector $y_{ij}={\Psi^{ws}}^T x_{ij}$, and then multiplied by a scaling function $\omega_k^{ws}=1/\tau^{ws}_k$ (whitening process). Truncating dimensions is equivalent to set $\omega_k^{ws}=0$ for these dimensions as done in Fisherface and many other variants of LDA \cite{Xiao1,Cevikalp,Chen1}. The scaling function is thus
\begin{equation} \label{eqn_2_FeatureWeighs}
\omega^{ws}_k=\left\{\begin{array}{ll}1/\sqrt{\lambda^{ws}_k},  & k\leq r_{ws}\\0,
& r_{ws}<k\leq l
\end{array}\right.,
\end{equation}
where $r_{ws} \leq min(l,\sum_{i=1}^C\sum_{j=1}^{H_i}(G_{ij}-1)),$ is the rank of $S_{ws}$. Similar scaling function ($\omega^w_k$) with $r_w \leq min(l,n-C)$ are also applicable for $S_w$ scatter matrix.

The inverses of $\tau_k^{ws}$ and $\tau_k^{w}$ pose two problems as shown in Fig. \ref{fig1_eigenValuesSwSws} right. Firstly, the eigenvectors corresponding to the zero eigenvalues are discarded or truncated as the features in the null space are weighted by a constant zero. This leads to the loses of important discriminative information that lies in the null space \cite{Cevikalp,Liu,Mandal1,Chen1}. Secondly, when the inverse of the square of the eigenvalues are used to scale the respective eigenvectors, features get undue weighage, noises get amplified and tend to over-fit the training samples.

\subsubsection{Within-subclass eigenspectrum modeling}

We use a median operator and its parameter similar to that used in \cite{Jiang4,Mandal2} to find the pivotal point $m$ for decreasing the decay of the eigenspectrum. A typical such $m$ value of a real eigenspectrum is shown in Fig. \ref{fig1_eigenValuesSwSws} right. We use the function form $1/f$, similar to \cite{Moghaddam4,Jiang4}, to estimate the eigenspectrum as $\tilde{\lambda}^{ws}_k=\frac{\alpha}{k+\beta},~~~~~1\leq k\leq r_{ws},$
where $\alpha$ and $\beta$ are two constants, used to model the real eigenspectrum in the initial portion. We determine $\alpha$ and $\beta$ by letting $\tilde{\lambda}^{ws}_1=\lambda^{ws}_1$ and $\tilde{\lambda}^{ws}_m=\lambda^{ws}_m$,
which yields $\alpha=\frac{\lambda^{ws}_1\lambda^{ws}_m(m-1)}{\lambda^{ws}_1-\lambda^{ws}_m}$,
$\beta=\frac{m\lambda^{ws}_m-\lambda^{ws}_1}{\lambda^{ws}_1-\lambda^{ws}_m}$.
Fig. \ref{fig1_eigenValuesSwSws} right, shows inverse of the square roots of a real eigenspectrum,
$\omega_k^{ws}=1/\tau^{ws}_k$, where $\tau^{ws}_k=\sqrt{\lambda^{ws}_k}$, and the stable portion of its model,
$\tilde{\omega}_k^{ws}=1/\tilde{\tau}^{ws}_k$, such that, $\tilde{\tau}^{ws}_k=\sqrt{\tilde{\lambda}^{ws}_k}$. We see that the model $\tilde{\omega}_k^{ws}$ fits closely to the real $\omega_k^{ws}$ in the reliable space but has slower decay in the unstable space.

\subsubsection{Feature extraction}
From Fig. \ref{fig1_eigenValuesSwSws} right, it is evident that noise component is small
as compared to face components in range space but it is dominating in unstable region. Thus, the estimated eigenspectrum $\tilde{\lambda}^{ws}_k$ is given by
\begin{equation}\label{eqn_9_EigenRegularization}
\tilde{\lambda}^{ws}_k=\left\{\begin{array}{ll}\lambda^{ws}_k,  & k< {m}\\
\frac{\alpha}{k+\beta},  & {m}\leq k\leq r_{ws}\\
\frac{\alpha}{r_{ws}+1+\beta}, & r_{ws}<k\leq l
\end{array}\right.
\end{equation}
The feature scaling function is then
$\tilde{\omega}^{ws}_k=\frac{1}{\sqrt{\tilde{\lambda}^{ws}_k}}, ~~ k=1,2,...,l.$
Fig. \ref{fig1_eigenValuesSwSws} right, shows the proposed feature scaling
function $\tilde{\omega}^{ws}_k$. Using this scaling function and
the eigenvectors $\psi^{ws}_k$, training data are transformed to
$\tilde{y}_{ij}=\mathbf{\tilde{\Psi}}_l^{{ws}T}x_{ij},$
where
$\mathbf{\tilde{\Psi}}_l^{ws}=[\tilde{\omega}^{ws}_k\psi^{ws}_k]_{k=1}^l$.
New between-subclass and total subclass scatter matrices are formed by vectors
$\tilde{y}_{ij}$ of the transformed training data as
\begin{equation}
\left\{
\begin{array}{lll}
\tilde{S}_{bs}=\sum^C_{i=1}\frac{p_i}{H_i}\sum^{H_i}_{j=1}(\tilde{\mu}_{ij}-\tilde{\mu})(\tilde{\mu}_{ij}-\tilde{\mu})^T,\\
\tilde{S}_{ts}=\sum_{i=1}^{C}\frac{p_i}{n_i}\sum_{j=1}^{n_i}(\tilde{y}_{ij}-\tilde{\mu})(\tilde{y}_{ij}-\tilde{\mu})^T,
\end{array}
\right.
\end{equation} where $\tilde{\mu}_{ij}=\frac{1}{G_{ij}}\sum_{k=1}^{G_{ij}}\tilde{y}_{ijk}$ and $\tilde{\mu}=\frac{1}{C}\sum_{i=1}^C\tilde{\mu}_i$, such that $\tilde{\mu}_i=\frac{1}{n_i}\sum_{j=1}^{n_i}\tilde{y}_{ij}$. In this work, we employ the total scatter matrix $\tilde{S}_{ts}$ of the regularized training data to extract the discriminative features because of its greater noise tolerance as compared to $\tilde{S}_{bs}$. The transformed features $\tilde{y}_{ij}$ will be de-correlated for $\tilde{S}_{ts}$ by solving the eigenvalue problem. Selecting the eigenvectors with the $d$ largest eigenvalues,
$\mathbf{\tilde{\Psi}}_d^{ts}=[\tilde{\psi}^{ts}_k]_{k=1}^d$, the proposed feature scaling and extraction matrix is given by $\mathbf{U}=\mathbf{\tilde{\Psi}}_l^{ws}\mathbf{\tilde{\Psi}}_d^{ts}$, which transforms a face image vector $x$, $x\in \mathbb{R}^{l}$, into a feature vector $z$, $z\in \mathbb{R}^{d}$, by $z=\mathbf{U}^Tx$.

\section{Experimental results}
Three popular benchmark databases are used to evaluate our proposed approach of whole space subclass discriminant analysis (WSSDA) for FR. All images are normalized following the CSU Face Identification Evaluation System \cite{Beveridge}. We follow the rule described in \cite{Zhu1} that every class is partitioned by the same number of subclasses $h$ (equally balanced), such that $H_i=h, \forall i$. $k$-means tree is built based on nearest neighbor (NN) clustering of face appearances, abbreviated as WSSDA-NN. We divide each class into two subclasses for AR and FERET databases and each class into four subclasses for YouTube database. We test our approach using PCA and RP (random projection) decision trees \cite{Wang4}, abbreviated as WSSDA-pcaTree and WSSDA-rpTree respectively. Cosine distance measure and the first nearest neighborhood classifier (1-NN) are applied to test the proposed WSSDA approach.

\subsection{Results on AR database}
\begin{figure}[!htp]
\centering
\includegraphics*[width=2.0in]{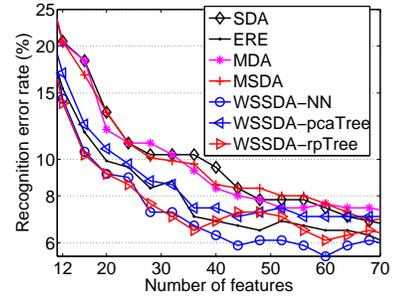}
\caption{Recognition error rate against the number of features used in the
matching on the AR database of 525 training images (75 people) and 525 testing
images (75 people).}
\label{fig3_AR75X7}
\end{figure}

In AR database \cite{Martinez2}, color images are converted to gray-scale and cropped into the size of $120\times170$ same as that in \cite{Martinez2,Jiang4}. Seventy-five subjects with 14 non-occluded images per subject are selected from the AR database. The first 7 images of 75 subjects are used in the training and also serve as gallery images. The second 7 images of the 75 subjects serve as probe images. Fig. \ref{fig3_AR75X7} shows the recognition error rate on the test set against the number of features $d$ used in the matching.
Methods that discard the null space of the within-class or within-subclass scatter matrices perform poorly as compared to the methods that utilize this valuable space for discriminant analysis. ERE approach evaluates the discriminant value in the whole space of $S_w$, so it performs better than all other approaches except our method. Partitioning each class into subclasses has helped in better capturing the variances arising from within-subclass scatter matrix. The proposed WSSDA approaches evaluate the discriminant value in the whole space of $S_{ws}$, hence they alleviate the over-fitting problem. Among them, WSSDA-NN outperform all other approaches across all different number of features.

\subsection{Results on FERET database}
\begin{figure}[!htp]
\centering
\includegraphics*[width=2.0in]{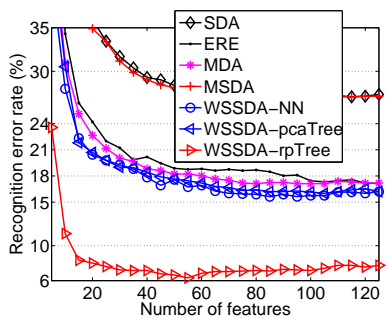}
\caption{Recognition error rate against the number of features used in the
matching on the FERET database of 512 training images (128 people) and 512
testing images (128 people).}
\label{fig4_FERET128X4Av}
\end{figure}

Using this database, we test the performance of the proposed algorithm on how well it generalizes to new test datasets (with new subjects), when trained with different set of subjects. It is constructed with normalized image size of $130\times150$, similar to one data set used in \cite{Lu5,Jiang4,Jiang3}, by choosing 256 subjects with at least four images per subject. 512 images of the first 128 subjects are used for training and the remaining 512 images of another 128 subjects serve as testing images. There is no overlap in subjects between the training and testing sets. For each subject, the $i^{th}$ image is chosen to form the gallery set and the remaining 3 images serve as the probe images to be identified. Fig. \ref{fig4_FERET128X4Av} shows the average recognition error rates over the 4 probe sets, each of which has a different gallery set. \vspace{-0.4cm}

As described previously, methods like MSDA and SDA perform very badly because they discard the null space of the within-class scatter matrix and their eigenfeatures are inversely weighted by very small or near zero eigenvalues causing severe over-fitting problem. MDA outperforms ERE, SDA and MSDA  approaches because it performs discriminant analysis using the scatter matrix arising from within-subclass variances. However, it discards the crucial null space of the within-subclass scatter matrix. The proposed WSSDA approaches (especially with rp-Tree partition) achieve consistently lowest recognition error rates for all number of features.

\subsection{Results on YouTube database}
YouTube faces (YTF) database \cite{Wolf1} contains 3,425 videos of 1,595 different people under real-world scenarios and hence very challenging. The longest and shortest videos contain 48 and 6070 frames, respectively, with an average of 181.3 frames per video. We directly crop the image centered on the face according to the provided data \cite{Wolf1} and then resize them into $40\times24$ which is similar to \cite{Cui1}.
\begin{figure}[!htp]
\centering
\includegraphics*[width=2.1in]{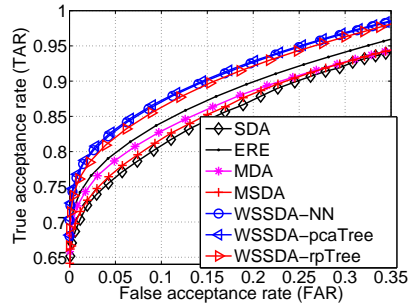}
\caption{TAR against FAR on YouTube database \cite{Wolf1}.}
\label{fig5_youTubeverifyTrg250X9Av}
\end{figure}
We use the same 5,000 video pairs from the YTF database described in \cite{Wolf1} and follow their splits/partitioning schemes for restricted video face verification. Experimental results are presented in this work for each approach where the minimum equal error rate is obtained. Fig. \ref{fig5_youTubeverifyTrg250X9Av} shows the average receiver operating characteristics (ROC) curves that plots the true acceptance rate (TAR) against the false acceptance rate (FAR) following the 10-fold cross-validation pairwise tests protocol suggested for the YTF database \cite{Wolf1}. It shows again that the proposed WSSDA approaches achieve higher TAR for their corresponding FAR among other tested approaches consistently for all different operating points.

For an accurate record, verification performance in terms of equal error rate (EER) obtained from the above experiments are shown in Table \ref{Tab_1}. From Fig. \ref{fig5_youTubeverifyTrg250X9Av} and Table \ref{Tab_1}, we observe that the proposed WSSDA approaches, {\em i.e.} WSSDA-NN, WSSDA-pcaTree and WSSDA-rpTree approaches outperform recent results on YTF database and achieves lowest EER among all the compared approaches except the DeepFace which uses an external data of 4.4 million face images for training \cite{Taigman1}. \vspace{-0.4cm}
\begin{table}[!htb]
    \caption{Equal Error Rate (EER \%) of various approaches on YouTube face image database.}
    \begin{minipage}{.485\linewidth}
      \label{Tab_1}
      \centering \footnotesize
        \begin{tabular}{|c|c|c|c|}
            \hline
            Method & ERR\\
            \hline
            MBGS \cite{Wolf1} & $25.3$\\
            \hline
            APEM Fusion \cite{Li3} & $21.4$\\
            \hline
            STFRD+PMML \cite{Cui1} & $19.9$\\
            \hline
            MMMRF Fusion \cite{Rowden1} & $12.6$\\
            \hline
            DeepFace \cite{Taigman1} & $8.6$\\
            \hline
        \end{tabular}
    \end{minipage}%
    \begin{minipage}{.55\linewidth}
      \centering \footnotesize
        \begin{tabular}{|c|c|c|c|}
            \hline
            Method & ERR\\
            \hline
            SDA & $14.0$\\
            \hline
            ERE & $12.8$\\
            \hline
            MDA & $13.1$\\
            \hline
            MSDA & $13.6$\\
            \hline
            \textbf{WSSDA-NN} & $\textbf{11.8}$\\
            \hline
            \textbf{WSSDA-pcaTree} & $\textbf{11.6}$\\
            \hline
            \textbf{WSSDA-rpTree} & $\textbf{12.2}$\\
            \hline
        \end{tabular}
    \end{minipage}
\end{table}

\vspace{-0.7cm}
\section{Conclusions}
This paper addresses the problems of discriminant analysis using within-class and within-subclass scatter matrices for FR. Each class is divided into subclasses using spatial partition trees so as to approximate the underlying distribution with mixture of Gaussians and perform whole space subclass discriminant analysis among these subclasses. This work proposes a regularization methodology that enables discriminant analysis in the whole eigenspace of the within-subclass scatter matrix. Low dimensional face discriminative features are extracted after performing discriminant evaluation in the entire eigenspace of within-subclass scatter matrix. Experimental results on popular databases, AR, FERET and challenging unconstrained YouTube face database show the superiority of our proposed approach on all three databases.


\bibliographystyle{IEEEbib}
\bibliography{face}

\end{document}